\documentclass{llncs}
\usepackage[cmex10]{amsmath}
\usepackage{algorithmic}
\usepackage{array}
\usepackage{mdwmath}
\usepackage{mdwtab}
\usepackage{eqparbox}
\usepackage[tight,footnotesize]{subfigure}
\usepackage{fixltx2e}
\usepackage{stfloats}
\usepackage{url}
\usepackage[utf8]{vietnam}
\usepackage[english]{babel}
\usepackage{adjustbox}
\usepackage{amsmath}
\usepackage{color}

\DeclareMathOperator*{\argmax}{arg\,max}

\begin{document}
\title{End-to-end Recurrent Neural Network Models for Vietnamese Named
  Entity Recognition: Word-level vs. Character-level} 

\author{Thai-Hoang Pham\inst{1} \and Phuong Le-Hong\inst{2}}
\institute{R\&D Department, Alt Inc, Hanoi, Vietnam\\
\email{phamthaihoang.hn@gmail.com},
\and
College of Science\\
Vietname National University in Hanoi, Vietnam\\
\email{phuonglh@vnu.edu.vn}}

\maketitle              

\begin{abstract}
  This paper demonstrates end-to-end neural network architectures for
  Vietnamese named entity recognition. Our best model is a
  combination of bidirectional Long Short-Term Memory (Bi-LSTM),
  Convolutional Neural Network (CNN), Conditional Random Field (CRF),
  using pre-trained word embeddings as input, which achieves an
  $F_{1}$ score of 88.59\% on a standard test set.  Our system is able
  to achieve a comparable performance to the first-rank system of the
  VLSP campaign without using any syntactic or hand-crafted
  features. We also give an extensive empirical study on using common
  deep learning models for Vietnamese NER, at both word and character
  level.  
\end{abstract}
\section{Introduction}

Named entity recognition (NER) is a fundamental task in natural
language processing and information extraction. It involves
identifying noun phrases and classifying each of them into a
predefined class. In 1995, the 6th Message Understanding Conference
(MUC)\footnote{\url{http://cs.nyu.edu/faculty/grishman/muc6.html}}
started evaluating NER systems for English, and in subsequent shared
tasks of CoNLL
2002\footnote{\url{http://www.cnts.ua.ac.be/conll2002/ner/}} and CoNLL
2003\footnote{\url{http://www.cnts.ua.ac.be/conll2003/ner/}}
conferences, language independent NER systems were evaluated. In these
evaluation tasks, four named entity types were considered, including
names of persons, organizations, locations, and names of miscellaneous
entities that do not belong to these three types.

More recently, the Vietnamese Language and Speech Processing
(VLSP)\footnote{\url{http://vlsp.org.vn/}} community has organized an
evaluation campaign to systematically compare NER systems for the
Vietnamese language. Similar to the CoNLL 2003 share task, four named
entity types are evaluated: persons (PER), organizations (ORG),
locations (LOC), and miscellaneous entities (MISC). The data are
collected from electronic newspapers published on the web.

In this paper, we present a state-of-the-art NER system for the
Vietnamese language without using any hand-crafted features. Our
system is competitive with the first-rank system of the VLSP campaign
that used many syntactic and hand-crafted features. In summary, the  
overall $F_1$ score of our system is 88.59\% on the standard test
set provided by the organizing
committee of the evaluation campaign\footnote{The
  first-rank system of the VLSP 2016 NER evaluation campaign has
  $F_1=$88.78\% on the test set.}. The contributions of this work include:
\begin{itemize}
\item We propose a truly end-to-end deep learning model which gives
  the state-of-the-art performance on a standard NER data set for
  Vietnamese. Our best model is a combination of Bi-LSTM, CNN, and
  CRF models, which achieves an $F_1$ score of 88.59\%.
\item We give an extensive empirical study on using common deep
  learning models for Vietnamese NER, at both word and character
  level. These models are also comparable to conventional sequence
  labeling models, including Maximum Entropy Markov Models (MEMMs) and
  CRFs.
\item We make our NER system open source for research purpose, which
  is believed to be a good contribution to the future development of
  Vietnamese NER in particular and Vietnamese language processing
  research in general.
\end{itemize}

The remainder of this paper is structured as
follows. Section~\ref{relatedWork} summarizes related work on
NER. Section~\ref{models} describes end-to-end models used in our
system. Section~\ref{experiments} gives experimental results and
discussions. Finally, Section~\ref{conclusion} concludes the paper.

\section{Related Work}
\label{relatedWork}
Within the large body of research on NER which have been published in
the last two decades, we identify two main approaches. The first
approach is characterized by the use of well-established sequence
labeling models such as conditional random field (CRF), hidden markov
model, support vector machine, maximum entropy and so on. The
performance of these models is heavily dependent on hand-crafted
features. In particular, most of the participants at CoNLL-2003 shared
task attempted to use information other than the available training
data such as gazetteers and unannotated data. The best system at
CoNLL-2003 shared task is the work of~\cite{Florian:2003} which
achieved an $F_{1}$ score of 88.76\%. After that,~\cite{Lin:2009}
surpassed them by using phrase features extracted from an external
database. Moreover, training NER models jointly with related tasks
helps improve their performance. For instance,~\cite{Durrett:2014}
trained a CRF model for joint-learning three tasks, including
coreference resolution, entity linking, and NER, and achieved the state-of-the-art result on OntoNotes dataset. With a similar
approach,~\cite{Luo:2015} gained the best performance on CoNLL-2003 shared
task dataset.

With a recent resurgence of the deep learning approach, several neural
architectures have been proposed for NER task. These methods have a long
story, but they have been focused only recently by the advance of
computational power and high-quality word embeddings. The first neural
network model is the work of~\cite{Petasis:2000} that used a
feed-forward neural network with one hidden layer. This model achieved
the state-of-the-art result on the MUC6 dataset. After
that,~\cite{Hammerton:2003} used a long short-term memory network for
this problem. Recently,~\cite{Collobert:2011} used a convolution
neural network over a sequence of word embeddings with a conditional
random field on the top. This model achieved near state-of-the-art
results on some sequence labeling tasks such as POS tagging,
chunking, and NER. From 2015 until now, the long short-term memory
model has been the best approach for many sequence labeling
tasks.~\cite{Huang:2015} used bidirectional LSTM with CRF layer for
joint decoding. Instead of using hand-crafted feature
as~\cite{Huang:2015},~\cite{Chiu:2016} proposed a hybrid model that
combined bidirectional LSTM with convolutional neural networks (CNN)
to learn both character-level and word-level
representations. Unlike~\cite{Chiu:2016},~\cite{Lample:2016} used
bidirectional LSTM to model both character and word-level
information. The work of~\cite{Ma:2016} proposed a truly end-to-end
model that used only word embeddings for detecting entities. This
model is the combination of CNN, bidirectional LSTM, and CRF
models. Approaching this problem at the character-level sequence, the
LSTM-CRF model of~\cite{Kuru:2016} achieved the nearly
state-of-the-art results in seven languages. 

\section{Methodology}
\label{models}
\subsection{Long Short-Term Memory Networks}

\subsubsection{Recurrent Neural Network}
The recurrent neural network (RNN) is a class of artificial neural
network designed for sequence labeling task. It takes input as a
sequence of vector and returns another sequence. The simple
architecture of RNN has an input layer $\textbf{x}$, hidden layer
$\textbf{h}$ and output layer $\textbf{y}$. At each time step $t$, the
values of each layer are computed as follows:
\begin{align*}
\textbf{h}_{t} &= f(\textbf{Ux}_{t}+\textbf{Wh}_{t-1})\\
\textbf{y}_{t} &= g(\textbf{Vh}_{t})
\end{align*}

where $\textbf{U}$, $\textbf{W}$, and $\textbf{V}$ are the connection
weight matrices in RNN, and $f(z)$ and $g(z)$ are sigmoid and softmax
activation functions.

\subsubsection{Long Short-Term Memory}
Long short-term memory (LSTM)~\cite{Hochreiter:1997} is a
variant of RNN which is designed to deal with these gradient vanishing and exploding
problems~\cite{Bengio:1994,Pascanu:2013} when learning with long-range sequences. LSTM networks are the same as RNN, except that the hidden
layer updates are replaced by memory cells. Basically, a memory cell
unit is composed of three multiplicative gates that control the
proportions of information to forget and to pass on to the next time
step. As a result, it is better for exploiting long-range dependency
data. The memory cell is computed as follows: 
\begin{align*}
\textbf{i}_{t}&=\sigma(\textbf{W}_{i}\textbf{h}_{t-1}+\textbf{U}_{i}\textbf{x}_{t}+\textbf{b}_{i})\\
\textbf{f}_{t}&=\sigma(\textbf{W}_{f}\textbf{h}_{t-1}+\textbf{U}_{f}\textbf{x}_t+\textbf{b}_{f})\\
\textbf{c}_{t}&=\textbf{f}_{t}\odot \textit{c}_{t-1}+\textbf{i}_{t}\odot\tanh(\textbf{W}_{c}\textbf{h}_{t-1}+\textbf{U}_{c}\textbf{x}_{t}+\textbf{b}_{c})\\
\textbf{o}_{t}&=\sigma(\textbf{W}_{o}\textbf{h}_{t-1}+\textbf{U}_{o}\textbf{x}_{t}+\textbf{b}_{o})\\
\textbf{h}_{t}&=\textbf{o}_{t}\odot\tanh(\textbf{c}_{t})
\end{align*}
where $\sigma$ is the element-wise sigmoid function and $\odot$ is the
element-wise product, $\textbf{i}$, $\textbf{f}$, $\textbf{o}$ and
$\textbf{c}$ are the input gate, forget gate, output gate and cell
vector respectively. $\textbf{U}_{i}, \textbf{U}_{f}, \textbf{U}_{c},
\textbf{U}_{o}$ are connection weight matrices between input
$\textbf{x}$ and gates, and $\textbf{U}_{i}, \textbf{U}_{f},
\textbf{U}_{c}, \textbf{U}_{o}$ are connection weight matrices between
gates and hidden state $\textbf{h}$. $\textbf{b}_{i}, \textbf{b}_{f},
\textbf{b}_{c}, \textbf{b}_{o}$ are the bias
vectors.

\subsubsection{Bidirectional Long Short-Term Memory}
The original LSTM uses only previous contexts for prediction. For many
sequence labeling tasks, it is advisable when taking the contexts
from two directions. Thus, we utilize the bidirectional LSTM
(Bi-LSTM)~\cite{Graves:2005,Graves:2013} for both word and
character-level systems.

\subsection{Conditional Random Field}
Conditional Random Field (CRF)~\cite{Lafferty:2001} is a type of
graphical model designed for labeling sequence of data. Although the
LSTM is likely to handle the sequence of the input data by learning
the dependencies between the input at each time step but it predicts
the outputs independently. The CRF, therefore, is beneficial to
explore the correlations between outputs and jointly decode the best
sequence of labels. In NER task, we implement the CRF on the top of
Bi-LSTM instead of the softmax layer and take outputs of Bi-LSTM as
the inputs of this model. The parameter of the CRF is the transition
matrix $A$ where $A_{i, j}$ represents the transition score from tag
$i$ to tag $j$. The score of the input sentence $\textbf{x}$ along
with the sequence of tags $\textbf{y}$ is computed as follow: 
\begin{equation*}
S(\textbf{x}, \textbf{y}, \theta \cup A_{i, j}) =
\sum_{t=1}^{T}(A_{{y}_{t-1}, {y}_{t}} + f_{{\theta}_{(y_{t},t)}}) 
\end{equation*}

where $\theta$ is the parameters of Bi-LSTM, $f_{\theta}$ is the score
outputed by Bi-LSTM, and $T$ is the number of time steps. Then the
tag-sequence likelihood is computed by the softmax equation: 
\begin{equation*}
p(\textbf{y}|\textbf{x}, A) = \frac{\exp(S(\textbf{x}, \textbf{y},
  \theta \cup A_{i, j}) )}{\sum_{\textbf{y}^{'} \in \textbf{Y}}
  \exp(S(\textbf{x}, \textbf{y}^{'}, \theta \cup A_{i, j}))} 
\end{equation*}
where $\mathbf{Y}$ is the set of all possible output sequences. In the
training stage, we maximize the log-likelihood function: 
\begin{equation*}
L = \sum_{i=1}^{N}\log p(\textbf{y}^{i}|\textbf{x}^{i};A)
\end{equation*}
where $N$ is the number of training samples. In the inference stage, the
Viterbi algorithm is used to find the output sequence $\textbf{y}^{*}$
that maximize the conditional probability: 
\begin{equation*}
\textbf{y}^{*} = \underset{\textbf{y} \in \textbf{Y}}{\argmax} \quad 
p(\textbf{y}|\textbf{x}, A) 
\end{equation*}

\subsection{Learning Word Embedings}

It has been shown that distributed representations of words
(words embeddings) help improve the accuracy of a various natural
language models. In this work, we investigate three methods to create 
word embeddings using a skip-gram model, a CNN model and a Bi-LSTM model.

\subsubsection{Pre-Trained Word Vectors Learnt by Skip-gram Model}

To create word embeddings for Vietnamese, we train a skip-gram model
using the word2vec\footnote{https://code.google.com/archive/p/word2vec/} tool on a dataset
consisting of 7.3GB of text from 2 million articles collected
through a Vietnamese news portal.\footnote{\url{http://www.baomoi.com}} 
The text is first normalized to lower case and all special characters are
removed. The common symbols such as the comma, the semicolon, the
colon, the full stop and the percentage sign are replaced with the special token \textit{punct}, and all numeral sequences
are replaced with the special token \textit{number}. Each word in the Vietnamese language may consist of more than one
syllables with spaces in between, which could be regarded as multiple
words by the unsupervised models. Hence it is necessary to replace the
spaces within each word with underscores to create full word
tokens. The tokenization process follows the method described
in~\cite{Le:2008a}. For words that appear in VLSP corpus but not appear
in word embeddings set, we create random vectors for these words by
uniformly sampling from the range
$[-\sqrt{\frac{3}{dim}},+\sqrt{\frac{3}{dim}}]$ where $dim$ is the
dimension of embeddings.

\subsubsection{Character-Level Word Vectors Learnt by Convolutional Neural Network}

\begin{figure}[t]
\centering
\includegraphics[scale=0.35]{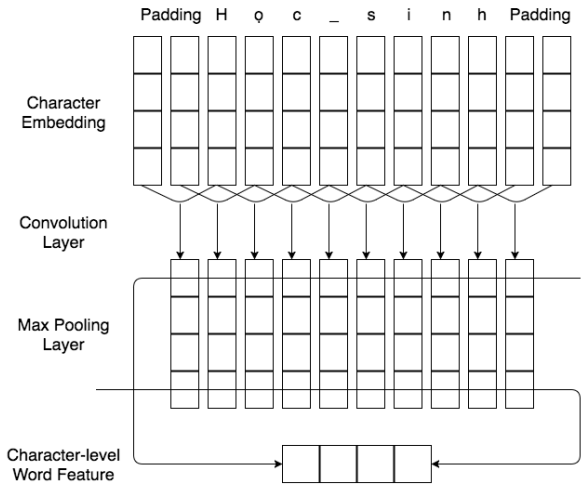}
\caption{The CNN for extracting character-level word features of word \textit{Học\_sinh} (Student).}
\label{fig:1}
\end{figure}

Convolutional neural network (CNN) is a type of feed-forward neural
networks that that uses many identical copies of the same neuron. This
characteristic of CNN permits this network to have lots of neurons
and, therefore, express computationally large models while keeping the
number of actual parameters relativity small. For NLP tasks, previous
works have shown that CNN is likely to extract morphological features
such as prefix and suffix effectively~\cite{Santos:2015,Chiu:2016,Ma:2016}. 
For this reason, we incorporate the CNN to the word-level
model to get richer information from character-level word vectors. These vectors are learnt
during training together with the parameters of the word models. The CNN we use in this paper is described in Figure~\ref{fig:1}. 

\subsubsection{Character-Level Word Vectors Learnt by Long Short-Term Memory}

\begin{figure}[t]
\centering
\includegraphics[scale=0.24]{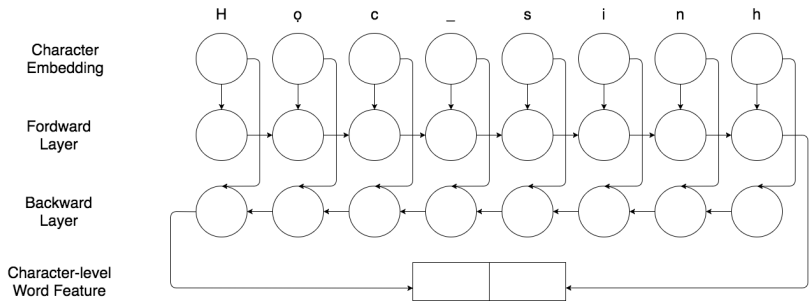}
\caption{The Bi-LSTM for extracting character-level word features of word \textit{Học\_sinh} (Student).}
\label{fig:4}
\end{figure}

The second way for generating character-level word vectors is using Bi-LSTM. In particular, we incorporate this model to the word-level
model to learn character-level word vectors. Character-level word vectors are concatenations of two last hidden states from forward and backward layers of Bi-LSTM. These vectors are also learnt
during training together with the parameters of the word models.
The Bi-LSTM model we use for this task is described in Figure~\ref{fig:4}. 

\subsection{Our Proposed Models}

We propose two different types of models based on the level of input,
either using word sequence or character sequence. Concretely, in the
first type, each input sentence is fed to the model as a sequence of
words, while in the second type, it is fed as a sequence of
characters. Both of the two model types share the same pipeline in
that it takes as input a sequence of distributed representations of
the underlying processing unit (word or character), that sequence is
then passed to a Bi-LSTM, and then a CRF layer takes as input the
output of the Bi-LSTM to predict the best named entity output
sequence.

\subsubsection{Word-Levels Models}

\begin{figure}[t]
\centering
\includegraphics[scale=0.22]{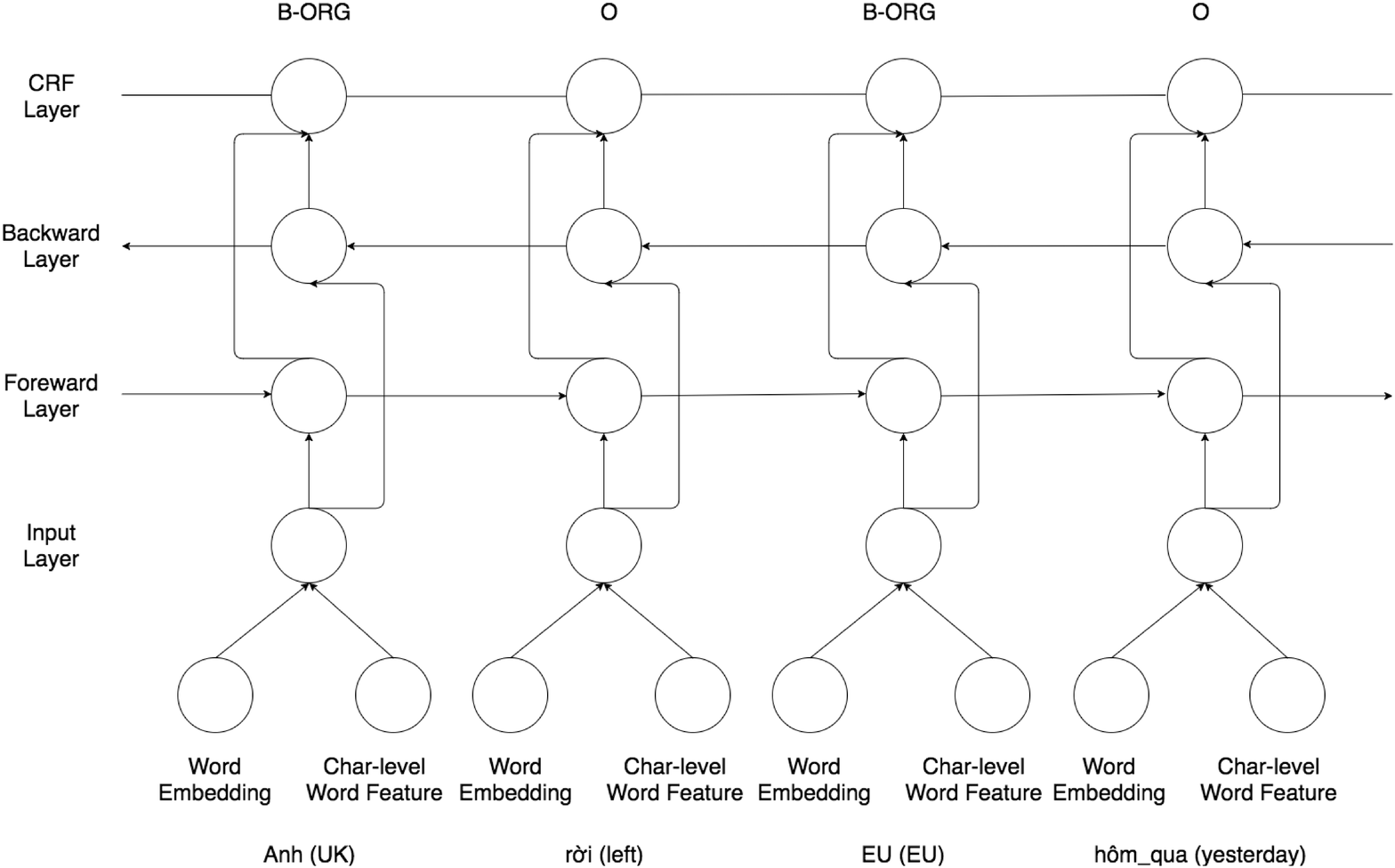}
\caption{Word-level model type for input sentence \textit{Anh rời EU hôm
    qua.} (UK left EU yesterday.) \textbf{Word-0} and \textbf{Word-1} models uses only word
  embeddings as input, while \textbf{Word-2} and \textbf{Word-3} models uses both word
  embeddings and word features generated either by CNN or Bi-LSTM.} 
\label{fig:2}
\end{figure}

In the first type, we investigate four different word embeddings,
including (\textbf{Word-0}) random vectors, (\textbf{Word-1})
skip-gram vectors, (\textbf{Word-2}) skip-gram vectors concatenated
with CNN-generated word features, and (\textbf{Word-3}) skip-gram
vectors concatenated with LSTM-generated word
features. Figure~\ref{fig:2} describes the architecture of the
word-level models.

\subsubsection{Character-Level Model}

In the second type, we investigate one model in that its input is a
sequence of vectors corresponding to characters of the input
sentence. We call this model (\textbf{Char-0}). Because the size of
Vietnamese character set is relatively small, our data set is
sufficient to learn distributed representations for Vietnamese
characters. We therefore initialize random vectors for these
characters by uniformly sampling from the range
$[-\sqrt{\frac{3}{dim}},+\sqrt{\frac{3}{dim}}]$ where $dim$ is the
dimension of embeddings. These character vectors are then learnt
during training together with the parameters of the models.

\begin{figure}[t]
\centering
\includegraphics[scale=0.22]{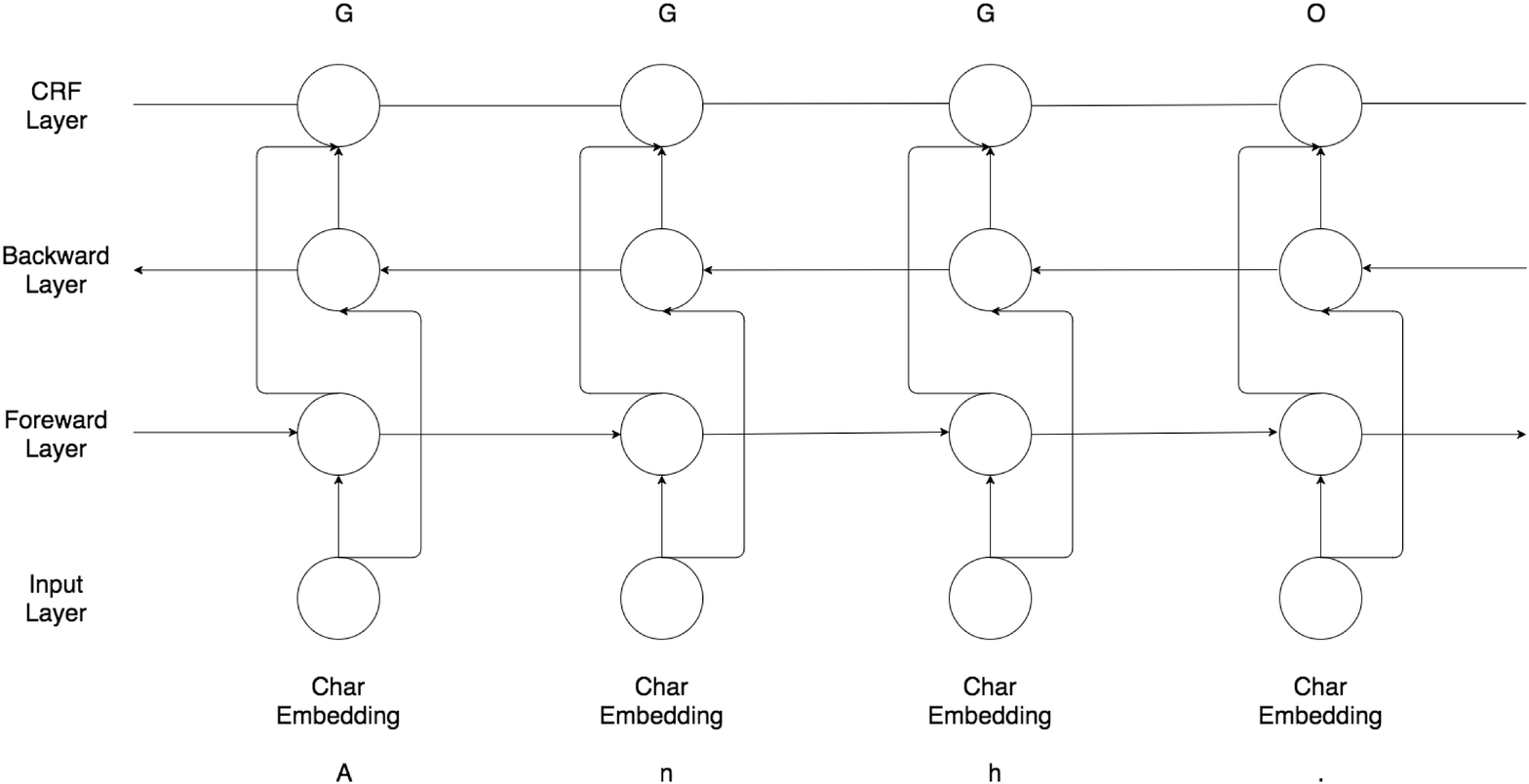}
\caption{Character-level model type for input sentence \textit{Anh.} (UK.)} 
\label{fig:5}
\end{figure}

The training data for NER is in CoNLL-2003 format, where both input
and output sequence are annotated at word-level. For this reason, it
is necessary to convert the dataset from word-level sequences to
character-level sequences. We use a simple method in which all
characters of a word are labeled with the same tag.  For example, the
label of all characters of a person named entity is P. Similarly, all
characters of location, organization, and miscellaneous tokens are
labelled with letters L, G, and M respectively. The characters of
other words and spaces are labelled by O. Figure~\ref{fig:3} shows the
transformation from word-level to character-level of an example
sentence \textit{Anh rời EU hôm qua} (UK left EU yesterday) and Figure~\ref{fig:5} describes the architecture of the
character-level models.
\begin{figure}
\centering
\begin{tabular}{cccc}
Anh & rời & EU & hôm\_qua \\ 
B-ORG & O & B-ORG & O \\ 
\end{tabular}\\
\vspace{0.2in}
\begin{adjustbox}{max width=0.5\textwidth}
\begin{tabular}{cccccccccccccccccc}
A & n & h &  & r & ờ & i &  & E & U &  & h & ô & m & \_ & q & u & a \\ 
G & G & G & O & O & O & O & O & G & G & O & O & O & O & O & O & O & O \\ 
\end{tabular} 
\end{adjustbox}
\caption{Word and character-level sequence labeling of the sentence \textit{Anh rời EU hôm\_qua.} (UK left EU yesterday.)}
\label{fig:3}
\end{figure}

\section{Results and Discussions}
\label{experiments}
\subsection{VLSP Corpus}
We evaluate our system on the VLSP NER shared task 2016 corpus. This
corpus consists of electronic newspapers published on the web. There are
four named entity types in this corpus, names of person, location,
organization and other named entities. Four types of NEs are compatible with their
descriptions in the CoNLL shared task 2003. The examples of each entity type are described in Table~\ref{tab:2} 

\begin{table}[h]
\caption{Examples of Vietnamese Entity Types\label{tab:2}}
\center
{
\begin{tabular}{|l|l|}
\hline 
Entity Types & Examples \\ 
\hline 
Person & \parbox[t]{2.1in}{thành phố Hồ Chí Minh (Ho Chi Minh city),
         núi Bà Đen (Ba Den mountain), sông Bạch Đằng (Bach Dang
         river)} \\  
\hline 
Location & \parbox[t]{2.1in}{công ty Formosa (Formosa company), nhà
           máy thủy điện Hòa Bình (Hoa Binh hydroelectric factory)} \\  
\hline 
Organization & \parbox[t]{2.1in}{ông Lân (Mr. Lan), bà Hà (Mrs. Ha)}\\ 
\hline 
Miscellaneous names & \parbox[t]{2.1in}{tiếng Indonesia (Indonesian),
                      người Canada (Canadian)} \\  
\hline 
\end{tabular}}
\end{table}

Data have been preprocessed with word segmentation and POS
tagging. Because POS tags and chunking tags are determined
automatically by public tools, they may contain mistakes. The format
of this corpus follows that of the CoNLL 2003 shared task. It consists
of five
columns. The order of these columns are word, POS tag, chunking tag,
named entity label, and nested named entity label. Our system focuses
on only named entity without nesting, so we do not use the fifth
column. Named entity labels are annotated using the IOB notation as in
the CoNLL shared tasks. There are 9 labels: B-PER and I-PER are used
for persons, B-ORG and I-ORG are used for organizations, B-LOC and
I-LOC are used for locations, B-MISC and I-MISC are used for other named entities 
and O is used for other elements. Table~\ref{tab:3} shows the quantity of named entity
annotated in the training set and the test set. 
\begin{table}[h]
\caption{Statistics of named entities in VLSP corpus\label{tab:3}}
\center
{
\begin{tabular}{|l|r|r|}
\hline 
Entity Types & Training Set & Testing Set \\ 
\hline 
Location & 6,247 & 1,379 \\ 
\hline 
Organization & 1,213 & 274 \\ 
\hline 
Person & 7,480 & 1,294 \\ 
\hline 
Miscellaneous names & 282 & 49 \\ 
\hline 
All & 15,222 & 2,996 \\ 
\hline 
\end{tabular}}
\end{table}

Because our systems are end-to-end architecture, we focus
only on the word and named entity label columns. To alleviate the data
sparseness, we perform the following preprocessing for our system: 
\begin{itemize}
\item{All tokens containing digit number are replaced by a special token \textit{number}.}
\item{All punctuations are replaced by a special token \textit{punct}.}
\end{itemize}

Moreover, we take one part of training data for
validation. The detail of each data set is described in
Table~\ref{tab:4}.

\begin{table}[t]
\caption{Size of each data set in VLSP corpus}
\center
\begin{tabular}{|l|r|}
\hline 
Data sets & Number of sentences \\ 
\hline 
Train & 14,861 \\ 
\hline 
Dev & 2,000 \\ 
\hline 
Test & 2,831 \\ 
\hline 
\end{tabular}
\label{tab:4} 
\end{table}

\subsection{Evaluation Method}
The performance is measured with $F_{1}$ score, where $F_1 =
\frac{2*P*R}{P + R}$. Precision ($P$) is the percentage of named entities found by
the learning system that are correct. Recall ($R$) is the percentage of
named entities present in the corpus that are found by the system. A
named entity is correct only if it is an exact match of the
corresponding entity in the data file. For character-level model,
after predicting label for each character, we convert these outputs
back to the word-level sequence to evaluate. The performance of 
our system is evaluated by the automatic
evaluation script of the CoNLL 2003 shared
task.\footnote{\url{http://www.cnts.ua.ac.be/conll2003/ner/}}.

\subsection{Results}

\subsubsection{Word-Level Model vs. Character-Level Model}

In the first experiment, we compare the effectiveness of word and
character-level approaches without using any external corpus. For this
reason, in this experiment, we do not use any pre-trained word
embeddings by comparing two models: \textbf{Word-0} and \textbf{Char-0}. Both of the two
models take embeddings as inputs of Bi-LSTM and predict outputs by
the CRF top layer. Table~\ref{tab:5} presents the performance of these systems.

\begin{table}[h]
\caption{Performances of word and character-level models}
\center
\begin{tabular}{|l|l|l|l|l|l|l|}
\hline 
Entity & \multicolumn{3}{c|}{\textbf{Word-0}} & \multicolumn{3}{c|}{\textbf{Char-0}}  \\ 
\hline 
 & $P$ & $R$ & $F_{1}$ & $P$ & $R$ & $F_{1}$ \\ 
\hline 
LOC & 88.37 & 74.69 & 80.95 & 80.03 & 84.84 & 82.37  \\ 
MISC & 90.48 & 77.55 & 83.52 & 84.21 & 65.31 & 73.56  \\ 
ORG & 60.57 & 38.83 & 47.32 & 50.00 & 33.58 & 40.17  \\ 
PER & 89.49 & 66.51 & 76.31 & 84.20 & 86.09 & 85.14  \\ 
\hline 
ALL & 86.78 & 67.90 & 76.19 & 80.08 & 80.37 & \textbf{80.23}  \\ 
\hline 
\end{tabular}
\label{tab:5}
\end{table}

We see that the character-level model outperforms the word-level model
by about 4\%. It is because the size of the character set is much smaller
than that of word set. The VLSP corpus, therefore, is enough for learning
effectively character embeddings. For word embeddings, we need a
bigger corpus to learn useful word vectors.

\subsubsection{Effect of Word Embeddings}

It is beneficial to use the external corpus to learn the word
embeddings. In the second experiment, we use skip-gram word
embeddings and compare \textbf{Word-1} and \textbf{Word-0} models. The
improvement by using pre-trained word embeddings for the word-level
model is shown in Table~\ref{tab:6}.

\begin{table}[h]
\caption{Performances of random and word2vec embeddings for word-level model}
\label{tab:6}
\center
\begin{tabular}{|l|l|l|l|l|l|l|}
\hline 
Entity & \multicolumn{3}{c|}{\textbf{Word-0}} & \multicolumn{3}{c|}{\textbf{Word-1}}  \\ 
\hline 
 & $P$ & $R$ & $F_{1}$ & $P$ & $R$ & $F_{1}$ \\ 
\hline 
LOC & 88.37 & 74.69 & 80.95 & 87.88 & 84.08 & 85.94  \\ 
MISC & 90.48 & 77.55 & 83.52 & 90.00 & 73.47 & 80.90  \\ 
ORG & 60.57 & 38.83 & 47.32 & 72.77 & 50.92 & 59.91  \\ 
PER & 89.49 & 66.51 & 76.31 & 88.92 & 71.38 & 79.19  \\ 
\hline 
ALL & 86.78 & 67.90 & 76.19 & 87.21 & 75.35 & \textbf{80.85}  \\ 
\hline 
\end{tabular}
\end{table}

By using pre-trained word embeddings, the performance of word-level
model increases by about 4\%, to 80.85\%. This accuracy is comparable to that of
the character-level model. It proves the effectiveness of using good
embeddings for both words and characters in the Bi-LSTM-CRF model. 

\subsubsection{Effect of Character-Level Word Features}

In the third experiment, we evaluate the performance of
\textbf{Word-2} and \textbf{Word-3} models. Recall that these two models 
make use of both pre-trained skip-gram word
embeddings and character-level word features generated either by CNN
or Bi-LSTM. The obtained performances are described in Table~\ref{tab:7}. 

\begin{table}[h]
\caption{Performances of word-level models}
\label{tab:7}
\center
\begin{tabular}{|l|l|l|l|l|l|l|l|l|l|}
\hline 
Entity & \multicolumn{3}{c|}{\textbf{Word-3}} & \multicolumn{3}{c|}{\textbf{Word-2}} & \multicolumn{3}{c|}{\textbf{Word-1}}  \\ 
\hline 
& $P$ & $R$ & $F_{1}$ & $P$ & $R$ & $F_{1}$ & $P$ & $R$ & $F_{1}$ \\ 
\hline 
LOC & 90.72 & 88.26 & 89.48&  91.60 & 88.85 & 90.20 & 87.88 & 84.08 & 85.94  \\ 
MISC & 94.29& 67.35& 78.57 & 97.30 & 73.47 & 83.72 & 90.00 & 73.47 & 80.90  \\ 
ORG & 69.23 & 52.75 & 59.88 &  72.77 & 62.64 & 67.32 & 72.77 & 50.92 & 59.91  \\ 
PER & 90.12 & 72.62 & 80.43 &  93.60 & 88.24 & 90.84 & 88.92 & 71.38 & 79.19  \\ 
\hline 
ALL &88.82 & 77.87 & 82.98 & 90.97 & 85.93 & \textbf{88.38} & 87.21 & 75.35 & 80.85  \\ 
\hline 
\end{tabular}
\end{table}

We observe a significant improvement of performance when
character-level word features learnt by CNN are integrated with pre-trained word
embeddings. This model achieves an overall $F_1$ score of 88.38\%. The
character-level word features learnt by Bi-LSTM are not as good as
those learnt by CNN, achieves only an overall $F_1$ score of 82.98\%,
but they also help improve the performance of the model in comparison
to the \textbf{Word-1} model.

\subsubsection{Comparison with Previous Systems}

In VLSP 2016 workshop, several different systems have been proposed
for Vietnamese NER. In this campaign, they have evaluated over three
entities types \textit{LOC}, \textit{ORG}, \textit{PER}. In all
fairness, we also evaluate our performances over these tags on the
same training and test set. The accuracy of our best model over three
entity types is 88.59\%, which is competitive with the best
participating system~\cite{Phuong:2016} in that shared task. That
system, however, used many hand-crafted features to improve the
performance of maximum entropy classifier (ME) while our system is
truly end-to-end model that takes only word sequences as inputs. Most
approaches in VLSP 2016 used the CRF and ME models, whose performance
is heavily dependent on feature engineering. Table~\ref{tab:8} shows
those models and their performance.

\begin{table}[h]
\caption{Comparison to participating NER systems at VLSP 2016}
\label{tab:8} 
\center

\begin{tabular}{|c|c|c|}
  \hline 
  \textbf{Team} & \textbf{Model} & \textbf{Performance} \\ 
  \hline
  \cite{Phuong:2016} & ME & 88.78  \\
  \hline
  \textbf{Word-2}&Bi-LSTM-CNN-CRF&88.59\\
  \hline 
  [Anonymous]\footnotemark & CRF & 86.62 \\ 
  \hline 
  \cite{Van:2016} & ME & 84.08 \\ 
  \hline
  \cite{Son:2016} & Bi-LSTM-CRF & 83.80 \\
  \hline 
  \cite{Huong:2016} & CRF  & 78.40 \\ 
  \hline 
\end{tabular}
\end{table}

\footnotetext{This team provided a system without the technical report.} 

There is one work~\cite{Son:2016} that applied deep learning approach
for this task. They used the implementation provided
by~\cite{Lample:2016}. There are two types of LSTM models in this open
source software: Bi-LSTM-CRF and Stack-LSTM. The model that is most
similar to ours is Bi-LSTM-CRF. The accuracy of this system is
83.25\%. Our system outperforms this model due to some possible
reasons. First, they used random vectors as word embeddings and update
them during the training stage. The VLSP corpus size is relatively
small so it is not good enough for learning word representations. Our
word embeddings are trained on a collection of Vietnamese newspapers
that is much larger and more abundant than the VLSP corpus. Second,
they used LSTM to model character-level features, while we used CNN in
our model. Previous works have shown that CNN is very useful to
extract these features~\cite{Santos:2015,Chiu:2016,Ma:2016}.

\section{Conclusion}
\label{conclusion}
In this work, we have investigated a variety of end-to-end recurrent
neural network architectures at both word and character-level for
Vietnamese named entity recognition. Our best end-to-end system
is the combination of Bi-LSTM, CNN, and CRF models, and uses
pre-trained word embeddings as input, which achieves an $F_1$ score
of 88.59\% on the standard test corpus published recently by the
Vietnamese Language and Speech community. Our system is competitive 
with the first-rank system of the related NER shared task without
using any hand-crafted features.

\section*{Acknowledgement}

The second author is partly funded by the Vietnam National University, Hanoi
(VNU) under project number QG.15.04.  Any opinions, findings and
conclusion expressed in this paper are those of the authors and do not
necessarily reflect the view of VNU.

%
%
\bibliographystyle{splncs03}
\bibliography{bibliography}
\end{document}